\documentclass[runningheads]{llncs}
\usepackage[T1]{fontenc}
\usepackage[utf8]{inputenc}
\usepackage{todonotes}
\usepackage{subcaption}
\usepackage{url}
\usepackage{graphicx}
\usepackage{amsmath}
\usepackage{amssymb}
\usepackage[hidelinks]{hyperref}
\usepackage{cleveref}
\usepackage{array}
\usepackage{booktabs}
\usepackage{multirow}
\usepackage{adjustbox}
\usepackage{footnote}
\usepackage{algorithm,algpseudocode}
\usepackage{xcolor,colortbl}
\usepackage[flushleft]{threeparttable}
\makesavenoteenv{tabular}
\makesavenoteenv{table}
\usepackage[misc]{ifsym}
\begin{document}
\title{Explaining Predictions by Characteristic Rules}
%
\author{Amr Alkhatib\orcidID{0000-0003-2745-6414}(\Letter) \and
Henrik Boström\orcidID{0000-0001-8382-0300} \and
Michalis Vazirgiannis\orcidID{0000-0001-5923-4440}}
\authorrunning{A. Alkhatib et al.}
%
\institute{KTH Royal Institute of Technology\\
Electrum 229, 164 40 Kista, Stockholm, Sweden\\ \email{\{amak2,bostromh,mvaz\}@kth.se}\\
}

\toctitle{Explaining Predictions by Characteristic Rules}
\tocauthor{Amr~Alkhatib,~Henrik~Boström,~Michalis~Vazirgiannis}
\maketitle              
\begin{abstract}

Characteristic rules have been advocated for their ability to improve interpretability over discriminative rules within the area of rule learning. However, the former type of rule has not yet been used by techniques for explaining predictions. A novel explanation technique, called CEGA (Characteristic Explanatory General Association rules), is proposed, which employs association rule mining to aggregate multiple explanations generated by any standard local explanation technique into a set of characteristic rules. An empirical investigation is presented, in which CEGA is compared to two state-of-the-art methods, Anchors and GLocalX, for producing local and aggregated explanations in the form of discriminative rules. The results suggest that the proposed approach provides a better trade-off between fidelity and complexity compared to the two state-of-the-art approaches; CEGA and Anchors significantly outperform GLocalX with respect to fidelity, while CEGA and GLocalX significantly outperform Anchors with respect to the number of generated rules. The effect of changing the format of the explanations of CEGA to discriminative rules and using LIME and SHAP as local explanation techniques instead of Anchors are also investigated. The results show that the characteristic explanatory rules still compete favorably with rules in the standard discriminative format. The results also indicate that using CEGA in combination with either SHAP or Anchors consistently leads to a higher fidelity compared to using LIME as the local explanation technique.

\keywords{Explainable machine learning \and Rule mining.}
\end{abstract}
\section{Introduction}\label{sec:the_ntro}
Machine learning algorithms that reach state-of-the-art performance, in domains such as medicine, biology, and chemistry~\cite{Pantelis-AI-review}, often produce non-transparent (black-box) models. However, understanding the rationale behind predictions is, in many domains, a prerequisite for the users placing trust in the models. This can be achieved by employing algorithms that produce interpretable (white-box)  models, such as decision trees and generalized linear models, but in many cases,  with a substantial loss of predictive performance~\cite{Loyola_8882211}. As a consequence, explainable machine learning has gained significant attention as a means to obtain transparency without sacrificing performance~\cite{molnar2019}.

Explanation techniques are either model-agnostic, i.e., they allow for explaining any underlying black-box model~\cite{Model:Agnostic}, or model-specific, i.e., they exploit properties of the underlying black-box model to generate the explanations, targeting e.g., random forests~\cite{BostroemGurungLindgren2018_1000117720,sirus-benard21a} or deep neural networks~\cite{deeplift-shrikumar17a,wangCNNExplainerLearning2020}. Along another dimension, the explanation techniques can be divided into local and global approaches~\cite{molnar2019}. Local approaches, such as LIME~\cite{lime} and SHAP~\cite{shap} aim to explain a single prediction of a black-box model~\cite{lime,shap}, while global approaches, such as SP-LIME~\cite{lime} and MAME~\cite{MAME}, aim to provide an understanding of how the model behaves in general~\cite{molnar2019}. Many explanation techniques produce explanations in the form of (additive) feature importance scores. Such explanations do however not directly lend themselves to verification, due to lack of an established, general and objective way of concluding whether the scores (or rankings imposed by them) are correct or not~\cite{advlime:aies20}. In contrast, some techniques, such as Anchors~\cite{anchors}, produce explanations in the form of rules. Since each such rule can be used to make predictions, the agreement (fidelity) of the rule to the underlying black-box model can be measured, e.g., using independent test instances. However, in some cases, the produced rules may be very specific~\cite{delaunay:hal-03133223}, which in practice precludes verification due to the limited coverage of the rules.

\begin{figure}[h]
    \centering
    \includegraphics[width=0.9\textwidth]{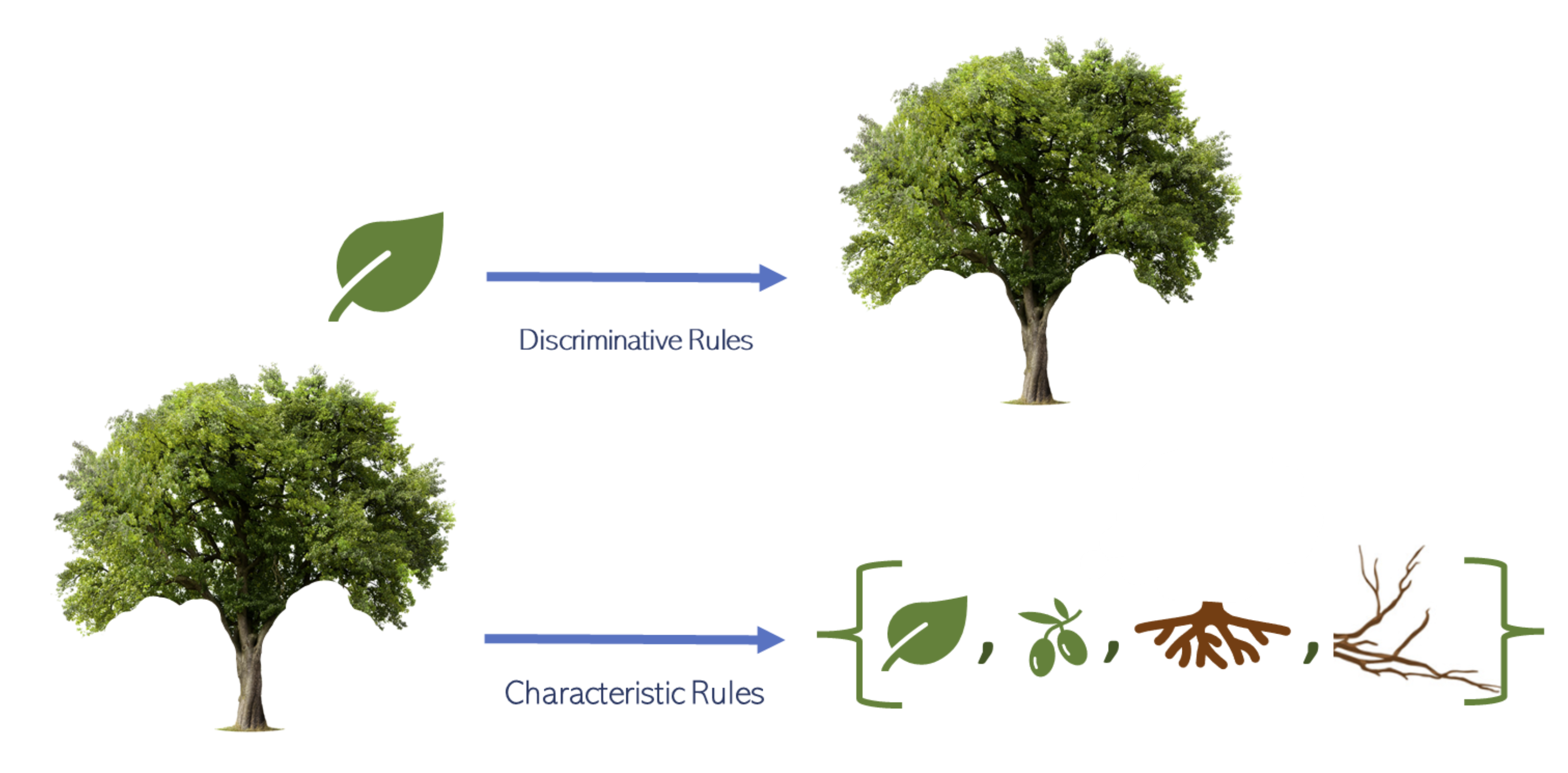}
    \caption{Discriminative rules distinguish one class from the others using a few features, while characteristic rules learn all the features that characterize the class.}
    \label{fig:illustration}
\end{figure}

Setzu et al. proposed GLocalX~\cite{GLocalX} as a solution to the above problem, by which multiple local explanations (rules) are merged into fewer, more general (global) rules. Similar to all local explanation techniques that output rules, GLocalX produces discriminative rules, which, according to Fürnkranz~\cite{Furnkranz}, provide a quick and easy way to distinguish one class from the others using a small number of features. Characteristic rules, on the other hand, capture properties that are common for objects belonging to a specific class, rather than highlighting the differences (only) between objects belonging to different classes. See Figure~\ref{fig:illustration} for an illustration of discriminative and characteristic rules. Although most rule learning approaches have targeted the former type of rule, also a few approaches for characteristic rule learning have been developed~\cite{Furnkranz,chr_rule_quan}.  As stated in~\cite[p.871]{Furnkranz}:

\begin{quote}
    ''Even though discriminative rules are easier to comprehend in the syntactic sense, we argue that characteristic rules are often more interpretable than discriminative rules.'' 
\end{quote}

Characteristic rules could hence be a potentially useful format also for explanations. Until now, however, the use of characteristic rules for explaining predictions have, to the best of our knowledge, not been considered. 

The main contributions of this study are:

\begin{itemize}
    \item a novel technique for explaining predictions by characteristic rules, called \textbf{C}haracteristic \textbf{E}xplanatory \textbf{G}eneral \textbf{A}ssociation rules (CEGA) 

    \item an empirical investigation comparing the fidelity and complexity of explanations in the form of characteristic rules, as produced by CEGA, and explanations in the form of discriminative rules, as generated by Anchors and GLocalX
    
    \item an ablation study of CEGA in which the format of the rules are changed from characteristic to standard discriminative rules and in which three different options for the local explanation technique are considered; LIME, SHAP and Anchors.
\end{itemize}

In the next section, we briefly discuss related work. In Section~\ref{CEGA}, we describe the proposed association rule mining approach for extracting general (and hence verifiable) explanations in the form of characteristic rules. In Section~\ref{Evaluation}, we present and discuss the results of a large-scale empirical investigation, in which the proposed method is compared to the baselines. Finally, in Section~\ref{CR}, we summarize the main findings and outline directions for future work. 

\section{Related Work}
\label{RW}

In this section, we start out by discussing model-agnostic explanation techniques that work with any algorithm for generating the black-box model. We then continue with model-specific explanation techniques that are explicitly designed for certain underlying models. We also cover some related work on rule learning.

Explainable machine learning is a research area that has gained quite some attention recently, in particular following the introduction of the popular local explanation technique LIME (Local Interpretable Model-agnostic Explanations)~\cite{lime}. In addition to the original algorithm, a variant called SP (Submodular Pick) LIME was proposed, which produces more general explanations. LIME trains a white-box model using perturbed instances, which are weighted by proximity to the instance of interest. The trained white-box model acts as a faithful explainer locally but not globally. On the other hand, SP-LIME uses a set of representative instances to generate explanations with high coverage, allowing for a general understanding of the black-box model. SHAP (SHapley Additive exPlanations)~\cite{shap} is another prominent, and computationally more efficient, local explanation technique based on game theory. It explains the outcome of a model by computing the additive marginal importance of each feature. The inventors of LIME subsequently proposed Anchors~\cite{anchors}, which generates explanations in the form of rules, such that non-included features should have no effect on the outcome of the model. It is claimed that the rules are easier to understand and often preferred to alternative types of explanation~\cite{anchors}. However, it has been shown that in some cases, Anchors may produce very specific (long) rules of low fidelity~\cite{delaunay:hal-03133223}. 
GLocalX~\cite{GLocalX} tries to overcome this problem by merging (similar) local explanations into a set of general rules. The generated (global) rules can be used to emulate the black-box model, in the sense that the rules can be used for making predictions.

In addition to the model-agnostic approaches, algorithm-specific methods for explaining predictions have also been proposed. An example of such a technique was proposed in~\cite{BostroemGurungLindgren2018_1000117720}, which generates explanations of random forest predictions by mining association rules from itemsets corresponding to paths in the trees. Similar approaches were proposed in~\cite{Deng2018InterpretingTE} and~\cite{sirus-benard21a}, however not primarily for explaining predictions, but for extracting (global) rule sets that approximate the underlying tree ensembles. The prominent RuleFit algorithm~\cite{FriedmanPopescu} also uses the prediction paths of the tree ensemble to form a global model, but instead of a rule set, a linear model is fit, using the Lasso to limit the complexity of the model. Again, the goal is not to explain predictions but to obtain a transparent and accurate predictive model. 

Highly related to techniques for explaining predictions by rules is the area of (inductive) rule learning~\cite{Furnkranz_rule_learning}. Early work in this area was presented in~\cite{MICHALSKI1983}, in which two different types of rule, discriminative and characteristic rules, were described. As described in~\cite{KLIEGR2021103458}, rules of the former type are on the form \verb! IF <conditions> THEN <class>!, where \verb!<conditions>! is a set of conditions, each referring to a feature, an operator and a value, and \verb!<class>! is a class label. They can differentiate one class from the other classes ''quickly'' according to Kliegr et al.~\cite{KLIEGR2021103458}. On the other hand, the characteristic rules have the opposite direction \verb! IF <class> THEN <conditions>!, which capture properties that are common for instances that belong to a certain class. The area of discriminative rule learning (or propositional rule induction) has been investigated to quite some extent, with contributions in the form of rule learning algorithms, such as CN2~\cite{Clark91ruleinduction}, RIPPER~\cite{Cohen95fasteffective}, and PRIM~\cite{Friedman1999}. In contrast, characteristic rule learning approaches are less common, one example being CaracteriX~\cite{chr_rule_quan}, which generates characteristic rules that can be applied to spatial and multi-relational databases.

\newpage

\section{From Local Explanations to General Characteristic Rules}
\label{CEGA}

In this section, we describe the proposed method to explain predictions by characteristic rules. We first outline details of the method and then show how we can easily change the format of the generated rules from characteristic to discriminative, allowing for directly investigating the effect of the rule format without changing any other parts of the algorithm.

\subsection{Explanation Mining and Rules Selection}\label{sec:rules-mining}

The proposed method (CEGA) takes as input a set of objects (instances), a black-box model, a local explainer, and minimum thresholds for the confidence and support of generated rules. The ouput of CEGA is a set of characteristic rules. The method is summarized in \cref{alg:CEGA}. Below we will discuss the different components in detail.

\begin{algorithm}
\small
\caption{CEGA}
\label{alg:CEGA}
\begin{algorithmic}[1]
\State $\bf{Input}$: A set of objects $X$, a local explainer $L$, a black-box model $B$; confidence threshold $c$; minimum rule support $s$; class labels $\{C_1, \ldots, C_k\}$\\

\State $E \gets$  {\it Generate-Explanation-Itemsets}$(X, L, B)$\\
\State $R \gets$  {\it Find-Association-Rules}$(E, s, c)$\\
\State $R \gets$  {\it Filter-Rules}$(R, \{C_1, \ldots, C_k\})$\\

\State $\bf{Output}$: General characteristic rules $(R)$
\end{algorithmic}
\end{algorithm}

The function {\it Generate-Explanation-Itemsets} uses the black-box model to obtain predictions (class labels) for the given objects and then employs the local explainer to obtain explanations for those predictions. The explanations are subsequently transformed into a list of itemsets, called {\it explanation itemsets}, where each itemset represents a local (instance-specific) explanation. Since CEGA is agnostic to the local explanation technique, the user may quite freely select the desired local explanation technique. However, different explanation techniques provide explanations in different formats, e.g., feature scores or rules. Consequently, each explanation format requires to be properly processed to produce the itemsets as required by CEGA. In case the local explanation technique outputs rules, such as Anchors, one processing option, which is the one that we will consider in this study, is to form an itemset from all the conditions of a rule together with the predicted class label. In the case that the local explanation technique outputs feature scores, such as LIME and SHAP, we will consider the option of forming itemsets by including the top-ranked features (in favor of the predicted class) together with the predicted class label. We binarized the categorical features using one-hot encoding, and the binarized feature names that appear in an explanation are added to the itemset (e.g., feature\_A\_V, where feature\_A is the name and V is the value), and the same preprocessing step is applied to the data at prediction time. Continuous features were discretized using equal-width binning, using five bins. As a side note, we extract more than one explanatory itemset per example, one per class, where each feature is added to an itemset with the class label it supports. Therefore, we obtain the same number of itemsets per class, which is particularly useful for highly imbalanced datasets to avoid extracting explanations solely for the dominant class. One final preprocessing step, for binary classification tasks only, is to add the binarized categorical feature to the itemset of the opposite class if the feature value is zero, which is motivated by the explainer associating the absence of one feature with the predicted class. Consequently, the presence of the same feature will favor the other class. This preprocessing step may hence result in that multiple values of a categorical feature may appear in characteristic explanations for the same class. In this study, we will consider binary classification problems only; therefore, the ranking of the features is with respect to the predicted class label. To avoid including features with negligible effect on a prediction, a threshold will be employed to filter out low-ranking features, which also reduces the computational cost when later performing association rule mining on the corresponding itemsets. Now the explanation itemsets can be used as input to the association rule mining algorithm.

An association rule mining algorithm is applied to the explanation itemsets within the {\it Find-Association-Rules} function, using the specified confidence and support thresholds. It should be noted that CEGA is again agnostic to which algorithm is used for conducting this step.

Using the {\it Filter-Rules} function, CEGA aims to find a set of rules that characterize each class. The characteristic rules can be obtained from the set of discovered association rules by keeping only the rules for which a single class label appear in the antecedent (condition part) and some set of conditions in the consequent. Moreover, to simplify the resulting set of rules, they are processed in the following way. If there are two rules that have the same class label, while the conditions of one of them is a subset of the conditions of the other rule, and if the former rule has higher confidence, then the latter rule is removed. This last step is motivated by that it is likely to reduce the complexity and the number of conditions of the resulting rules while increasing the coverage of the resulting rule set. As discussed in~\cite{Furnkranz}, this may however not necessarily lead to that the resulting rules are indeed more interpretable.

\subsection{Discriminative vs. Characteristic Rules}\label{sec:chr-rules}

The confidence of a discriminative rule, where the antecedent consists of conditions and the consequent consists of a class label, is the probability that an object belongs to the class given that it satisfies the conditions~\cite{Agrawal:1994} (equation~\ref{eq1}). The confidence of a characteristic rule is conversely the probability that an object satisfies the conditions given that it belongs to the class (equation~\ref{eq2}). It should be noted that by just a slight modification of the function {\it Filter-Rules}, concerning whether the conditions and the label should appear in the antecedent and consequent (or vice versa), CEGA will produce discriminative instead of characteristic rules.

\small
\begin{equation}
\label{eq1}
    confidence({Conditions} \rightarrow {Class}) = P(Class | Conditions) 
    = \frac{P(Conditions, Class)}{P(Conditions)}
\end{equation}

\small
\begin{equation}
\label{eq2}
    confidence(Class \rightarrow Conditions) = P(Conditions | Class) 
    = \frac{P(Conditions, Class)}{P(Class)}
\end{equation}

The confidence in (equation~\ref{eq2}) for a characteristic rule measures how characteristic is a set of features given a class. For example, a rule with 100\% confidence means that the items (features) are shared between all class objects.

\section{Empirical Evaluation}
\label{Evaluation}

In this section, we compare CEGA to Anchors and GLocalX, two state-of-the-art approaches for explaining predictions with discriminative rules. The methods will be compared with respect to fidelity (agreement with the explained black-box model) and complexity (number of rules). We then conduct an ablation study, in which we change the rule format of CEGA from characteristic to discriminative rules (as explained in the previous section) and also the technique used to generate the local explanations.

\subsection{Experimental Setup}

The quality of the generated explanations will be estimated by their fidelity to the underlying black-box model. The fidelity can however be measured in different ways. Guidotti et al.~\cite{Guidotti_survey} define fidelity as the ability of an interpretable model to replicate the output of the black box model, as measured using some predictive performance metric, such as accuracy, F1 score, and recall. We follow this definition and provide a comprehensive set of fidelity measurements, including accuracy, F1-score, and area under the ROC curve (AUC), to report how well an explanation technique approximates the black-box model.

This means that for each object from which we have obtained a black-box prediction, we need to form also a prediction from the rules that are used to explain the prediction. We employ a naive Bayes strategy with Laplace correction~\cite{Kohavi97} to obtain predictions for both discriminative and characteristic rules.

The experiments have been conducted using 20 public datasets\footnote{All the datasets were obtained from \url{https://www.openml.org} except Adult, German credit, and Compas}. Each dataset is split into training, development, and testing sets, where the black-box model is trained on the first, the explanations (rules) are generated using the black-box predictions on the second set, and finally, the quality of the produced rules is evaluated on the third set. All datasets concern binary classification tasks except for Compas\footnote{\url{https://github.com/propublica/compas-analysis}}, which originally contains three classes (Low, Medium, and High), which were reduced into two by merging Low and Medium into one class. The black-box models are generated by XGBoost~\cite{xgboost}. Some of its hyperparameters (learning rate, number of estimators and the regularization parameter lambda) were tuned by grid search using 5-fold cross-validation on the training set. CEGA requires two additional hyperparameters; support and confidence. The former was set to 10 in the case of LIME and SHAP and set to 4 in the case of Anchors, while the confidence of the characteristic rules was tuned based on the fidelity as measured on the development set. In the second experiment, where CEGA was used to produce 
also discriminative rules, the confidence was set to 100\% for this rule type to keep the number of generated rules within a reasonable limit.

In the first experiment, Anchors, GLocalX and CEGA are compared, where the two latter use Anchors at the local explanation technique. Anchors is used with the default hyperparameters, and the confidence threshold has been set to 0.9. GLocalX is tested as well using the default values except for the alpha hyperparameter, in which values between 50 and 95 have been tested with step 5, and the best result is reported. In a second experiment, we consider using also SHAP and LIME as local explanation techniques for CEGA. As described in Section~\ref{sec:rules-mining}, the output of these techniques require some preprocessing to turn them into itemsets; the threshold to exclude low-ranking features has here been set to 0.01. In all experiments, CEGA will employ the Apriori algorithm~\cite{Agrawal:1994} for association rule mining\footnote{CEGA is available at: \url{https://github.com/amrmalkhatib/CEGA}}. 

\subsection{Baseline Experiments}\label{sec:baseline}

The results from comparing the characteristic rules of CEGA to GLocalX and Anchors are summarized in Table~\ref{tab:gloX}. It can be seen that CEGA performs on par with Anchors with respect to fidelity, while producing much fewer unique rules. At the same time, CEGA generally obtains a higher fidelity than GLocalX.

\begin{figure}[h]
    \centering
    \includegraphics[width=0.9\textwidth]{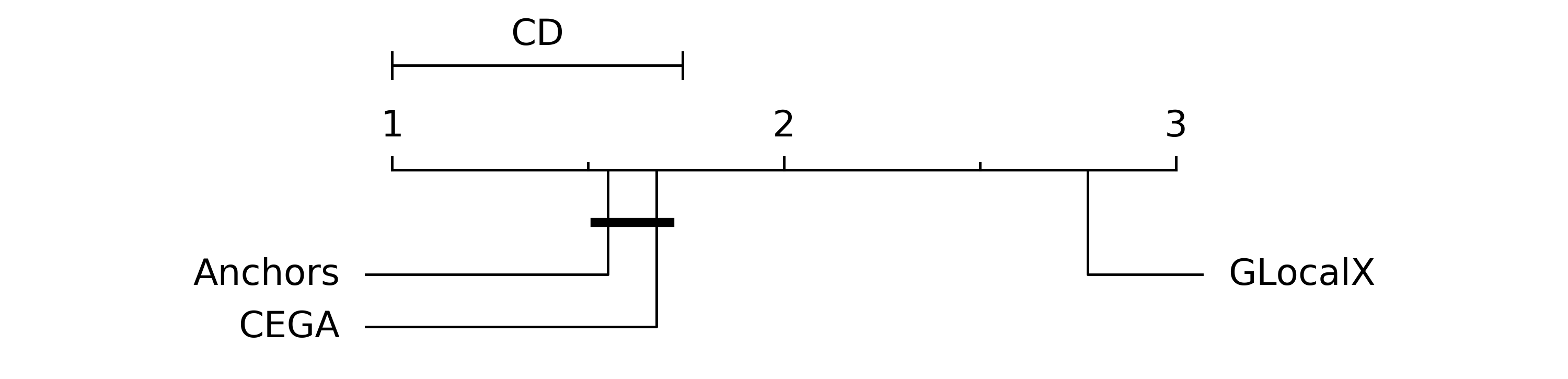}
    \caption{Comparing average ranks with respect to fidelity measured by AUC (lower rank is better) of Anchors, GLocalX, and CEGA, where the critical difference (CD) represents the largest difference that is not statistically significant.}
    \label{fig:auc_ranks}
\end{figure}

\begin{figure}[h]
    \centering
    \includegraphics[width=0.9\textwidth]{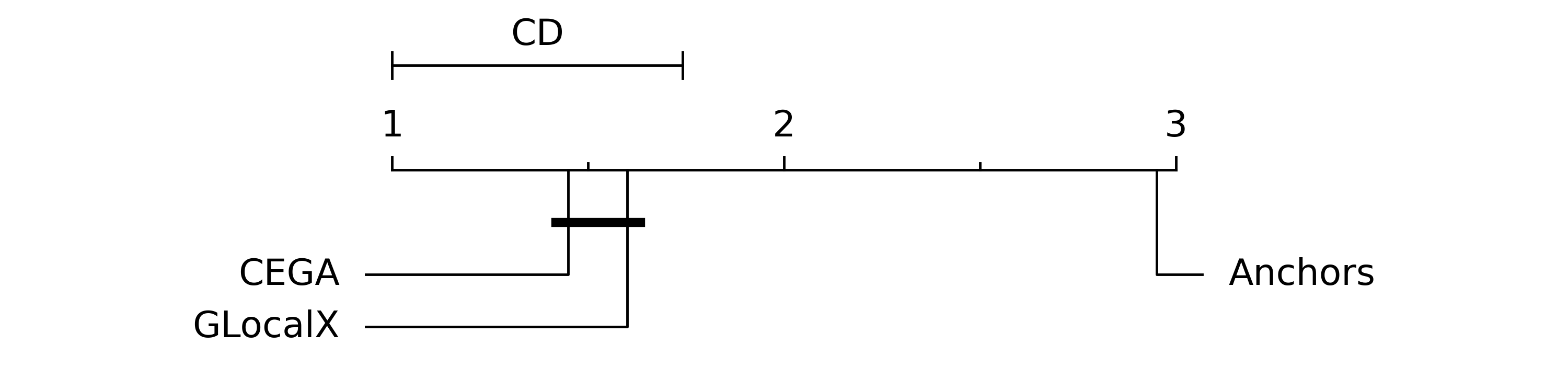}
    \caption{Comparing average ranks of Anchors, GLocalX, and CEGA with respect to the number of rules}
    \label{fig:num_rules_ranks}
\end{figure}

To test the null hypothesis that there is no difference in fidelity, as measured by AUC, between Anchors, GLocalX, and CEGA, the Friedman test~\cite{Friedman_test} followed by pairwise posthoc Nemenyi tests~\cite{nemenyi1963distribution}  are employed. The first test rejects the null hypothesis and the result of the post hoc tests are summarized in Fig.~\ref{fig:auc_ranks}. It can be observed that GLocalX is significantly outperformed with respect to fidelity by both Anchors and CEGA. In Fig.~\ref{fig:num_rules_ranks}, the result of the same test is shown when comparing the number of unique rules produced by the three approaches. This time, Anchors is significantly outperformed (produces more rules) than the two other approaches. In summary, it can be concluded that with CEGA, we can keep a level of fidelity that is not significantly different from using Anchors, while reducing the number of rules significantly; from hundreds of rules to just a handful. 

\begin{table}
\caption{Fidelity, number of rules and coverage for Anchors, GLocalX and CEGA.}
\begin{center}
\begin{adjustbox}{width=1.\textwidth}
\small
\begin{threeparttable}
\begin{tabular}{l c c c c c c@{\hskip 0.125in}c@{\hskip 0.125in}c c c c c c @{\hskip 0.125in}c @{\hskip 0.125in}c c c c c}
    \toprule
    \multicolumn{1}{c}{\multirow{2}{*}{\bfseries Dataset}} & 
    \multicolumn{5}{c}{\bfseries Anchors} &
    \multicolumn{9}{c}{\bfseries GLocalX} &
    \multicolumn{5}{c}{\bfseries CEGA}\\ \cmidrule(lr){2-20}
    \multicolumn{1}{c}{}& Acc.&AUC&F1&\#Rules\tnote{*}&Cov.\tnote{**}&&&Acc&AUC&F1&\#Rules&Cov.&&&Acc&AUC&F1&\#Rules&Cov. \\
    \cmidrule(lr){1-20}
    ada & 0.90 & \bf0.92 & 0.86 & 120 & 1.0 &&& 0.84 & 0.86 & 0.8 & 11 & 0.33 &&& 0.87 & 0.91 & 0.81 & 5 & 1.0\\ \cmidrule(lr){1-20}
    Adult\tnote{***} & 0.91 & \bf0.93 & 0.85 & 378 & 1.0 &&& 0.85 & 0.87 & 0.81 & 23 & 0.3 &&& 0.88 & 0.91 & 0.82 & 7 & 1.0\\ \cmidrule(lr){1-20}
    Bank Marketing & 0.93 & 0.67 & 0.48 & 88 & 1.0 &&& 0.94 & 0.61 & 0.65 & 57 & 0.39 &&& 0.91 & \bf0.80 & 0.65 & 11 & 1.0\\ \cmidrule(lr){1-20}
    Blood Transfusion & 0.91 & \bf0.97 & 0.91 & 15 & 1.0 &&& 0.95 & \bf0.97 & 0.9 & 3 & 0.19 &&& 0.93 & 0.92 & 0.86 & 4 & 0.92\\ \cmidrule(lr){1-20}
    BNG breast-w & 0.98 & 0.99 & 0.98 & 96 & 1.0 &&& 0.67 & 0.5 & 0.4 & 2 & 0.0 &&& 0.97 & \bf1.00 & 0.97 & 8 & 0.99\\ \cmidrule(lr){1-20}
    BNG tic-tac-toe & 0.84 & \bf0.87 & 0.79 & 842 & 0.998 &&& 0.28 & 0.33 & 0.28 & 158 & 0.33 &&& 0.75 & 0.77 & 0.72 & 4 & 1.0\\ \cmidrule(lr){1-20}
    Compas & 0.89 & \bf0.83 & 0.73 & 168 & 1.0 &&& 0.9 & 0.76 & 0.74 & 78 & 0.72 &&& 0.86 & 0.82 & 0.67 & 10 & 1.0\\ \cmidrule(lr){1-20}
    Churn & 0.89 & 0.67 & 0.47 & 143 & 1.0 &&& 0.88 & \bf0.71 & 0.69 & 80 & 0.34 &&& 0.81 & 0.66 & 0.54 & 20 & 1.0\\ \cmidrule(lr){1-20}
    German Credit\tnote{****} & 0.78 & 0.75 & 0.58 & 39 & 1.0 &&& 0.43 & 0.59 & 0.43 & 25 & 0.43 &&& 0.79 & \bf0.79 & 0.72 & 6 & 1.0\\ \cmidrule(lr){1-20}
    Internet Advertisements & 0.91 & 0.78 & 0.78 & 191 & 1.0 &&& 0.87 & 0.5 & 0.46 & 16 & 0.57 &&& 0.91 & \bf0.80 & 0.77 & 4 & 1.0\\ \cmidrule(lr){1-20}
    Jungle Chess 2pcs & 1.0 & \bf1.0 & 1.0 & 24 & 1.0 &&& 0.45 & 0.5 & 0.31 & 1 & 0.51 &&& 0.89 & 0.91 & 0.89 & 6 & 1.0\\ \cmidrule(lr){1-20}
    kc1 & 0.89 & \bf0.86 & 0.72 & 111 & 1.0 &&& 0.87 & 0.79 & 0.68 & 27 & 1.0 &&& 0.85 & \bf0.86 & 0.72 & 4 & 0.99\\ \cmidrule(lr){1-20}
    mc1 & 0.96 & 0.69 & 0.53 & 9 & 0.82 &&& 0.97 & 0.7 & 0.53 & 2 & 0.03 &&& 0.97 & \bf0.95 & 0.54 & 19 & 0.99\\ \cmidrule(lr){1-20}
    mofn-3-7-10 & 0.89 & 0.96 & 0.79 & 43 & 1.0 &&& 0.64 & 0.75 & 0.62 & 16 & 0.7 &&& 1.0 & \bf1.0 & 1.0 & 7 & 0.99\\ \cmidrule(lr){1-20}
    Mushroom & 0.99 & \bf1.0 & 0.99 & 89 & 1.0 &&& 0.83 & 0.84 & 0.83 & 3 & 0.63 &&& 0.88 & 0.93 & 0.88 & 3 & 1.0\\ \cmidrule(lr){1-20}
    Phishing Websites & 0.93 & \bf0.98 & 0.93 & 107 & 1.0 &&& 0.42 & 0.5 & 0.3 & 3 & 0.53 &&& 0.92 & 0.96 & 0.92 & 5 & 1.0\\ \cmidrule(lr){1-20}
    socmob & 0.95 & 0.94 & 0.91 & 36 & 1.0 &&& 0.18 & 0.5 & 0.15 & 3 & 0.04 &&& 0.96 & \bf0.99 & 0.93 & 7 & 1.0\\ \cmidrule(lr){1-20}
    Spambase & 0.94 & \bf0.98 & 0.93 & 193 & 0.998 &&& 0.6 & 0.5 & 0.37 & 4 & 0.41 &&& 0.90 & 0.97 & 0.90 & 8 & 0.93\\ \cmidrule(lr){1-20}
    Steel Plates Fault & 0.81 & \bf1.00 & 0.81 & 71 & 1.0 &&& 0.72 & 0.77 & 0.72 & 4 & 0.6 &&& 0.77 & 0.82 & 0.77 & 4 & 1.0\\ \cmidrule(lr){1-20}
    Telco Customer Churn & 0.89 & 0.93 & 0.84 & 167 & 1.0 &&& 0.82 & 0.86 & 0.79 & 24 & 0.37 &&& 0.89 & \bf0.95 & 0.85 & 8 & 1.0\\ \cmidrule(lr){1-20}
    Average rank\tnote{*****} & 1.5 & 1.55 & 1.55 & 2.95 & 1.425 &&& 2.525 & 2.775 & 2.65 & 1.6 & 2.925 &&& 1.975 & 1.675 & 1.8 & 1.45 & 1.65\\
    \bottomrule
\end{tabular}
\begin{tablenotes}
  \item[*] The number of rules
  \item[**] The coverage is the percentage of instances in the dataset that are covered by at least one rule
  \item[***] Available at: \url{https://archive.ics.uci.edu/ml/datasets/Adult}
  \item[****] Available at: \url{https://archive.ics.uci.edu/ml/datasets/statlog+(german+credit+data)}
  \item[*****] The average rank shows which method is better on average, with 1 being the best and 3 the worst result
  \end{tablenotes}
  \end{threeparttable}
\end{adjustbox}
\end{center}
\label{tab:gloX}
\end{table}

\newpage
\subsection{Comparing Discriminative and Characteristic Rules}\label{sec:evaluat}

Since CEGA allows for changing the rule format with just a minor modification to the algorithm, as described above, we can easily investigate whether CEGA's improved trade-off between fidelity and complexity (measured by the number of rules) compared to Anchors and GLocalX is due to the use of characteristic rules, or potentially comes from the other components, e.g., the use of association rule mining to generalize the rules. We will in this section present results from such a comparison. We will first however illustrate the differences between characteristic and discriminative explanatory rules. To this end, samples of  characteristic and discriminative rules are shown in Table~\ref{tab:chr_rules_examples} and Table~\ref{tab:disc_rules_examples}, respectively. In this study, we are not drawing any conclusions on their relative interpretability, as judged by human users, but will focus mainly on fidelity and complexity. 
In Table~\ref{tab:rulesAcc}, the fidelity, measured by accuracy, AUC and F1 score, of CEGA producing characteristic and discriminative rules is presented, together with the number of rules and the coverage.
Due to its computational efficiency, SHAP is here used as a local explanation technique for both versions of CEGA (instead of Anchors in the baseline experiments~\ref{sec:baseline}). One can observe that the fidelity of the characteristic rules tends to be higher than for the discriminative rules, while the characteristic rules are clearly fewer; in other words, very similar to what we could observe when comparing (standard) CEGA to Anchors and GLocalX. Since we only compare two methods in this experiment, the Wilcoxon signed-rank test~\cite{wilcoxon1945individual} is employed instead of the Friedman test to investigate if the observed difference in fidelity, as measured by AUC, is significant or not. It turns out that the null hypothesis may be rejected at the 0.05 level also here. 
 
\begin{table}[]
\caption{The top 11 characteristic rules output by CEGA for the German Credit dataset when using SHAP as the local explainer.}
\begin{center}
\begin{adjustbox}{width=0.8\textwidth}
\begin{tabular}{clc}
\hline

Label & \multicolumn{1}{c}{Conditions} & Confidence \\ \hline
\rowcolor[HTML]{DBE5F5} 
good  & credit\_history = critical/other existing credit       & 1.0        \\
\rowcolor[HTML]{FFCCC9} 
bad   & credit\_history = no credits/all paid                  & 1.0        \\
\rowcolor[HTML]{DBE5F5} 
good  & personal\_status = male single                         & 0.993      \\
\rowcolor[HTML]{DBE5F5} 
good  & purpose = used car                                    & 0.993      \\
\rowcolor[HTML]{FFCCC9} 
bad   & property\_magnitude = no known property                & 0.993      \\
\rowcolor[HTML]{FFCCC9} 
bad   & purpose = education                                   & 0.993      \\
\rowcolor[HTML]{DBE5F5} 
good  & other\_parties = guarantor                             & 0.987      \\
\rowcolor[HTML]{DBE5F5} 
good  & other\_payment\_plans = none                            & 0.987      \\
\rowcolor[HTML]{FFCCC9} 
bad   & purpose = new car                                     & 0.987      \\
\rowcolor[HTML]{DBE5F5} 
good  & checking\_status = no checking                         & 0.98       \\
\rowcolor[HTML]{FFCCC9} 
bad   & credit\_history = all paid                             & 0.953      \\ \hline
\end{tabular}
\end{adjustbox}
\end{center}
\label{tab:chr_rules_examples}
\end{table}

\begin{table}[]
\caption{The top 11 discriminative rules output by CEGA for the German Credit dataset when using SHAP as the local explainer.}
\begin{center}
\begin{adjustbox}{width=1.\textwidth}
\begin{tabular}{@{}lcc@{}}
\toprule
Conditions                                                           & Label & Confidence \\ \midrule
\rowcolor[HTML]{DBE5F5} 
other\_payment\_plans = none \& checking\_status = no checking           & good  & 1.0        \\
\rowcolor[HTML]{DBE5F5} 
other\_payment\_plans = none \& purpose = radio/tv                      & good  & 1.0        \\
\rowcolor[HTML]{DBE5F5} 
checking\_status = no checking \& purpose = radio/tv                   & good  & 1.0        \\
\rowcolor[HTML]{DBE5F5} 
other\_payment\_plans = none \& housing = own                           & good  & 1.0        \\
\rowcolor[HTML]{DBE5F5} 
checking\_status = no checking \& housing = own                        & good  & 1.0        \\
\rowcolor[HTML]{DBE5F5} 
property\_magnitude = real estate \& checking\_status = no checking     & good  & 1.0        \\
\rowcolor[HTML]{DBE5F5} 
property\_magnitude = real estate \& other\_payment\_plans = none        & good  & 1.0        \\
\rowcolor[HTML]{DBE5F5} 
housing = own \& purpose = radio/tv                                   & good  & 1.0        \\
\rowcolor[HTML]{DBE5F5} 
property\_magnitude = real estate \& housing = own                     & good  & 1.0        \\
\rowcolor[HTML]{DBE5F5} 
other\_payment\_plans = none \& savings status no known savings       & good  & 1.0        \\
\rowcolor[HTML]{FFCCC9} 
property\_magnitude = life insurance \& credit\_history = existing paid & bad   & 1.0        \\ \bottomrule
\end{tabular}
\end{adjustbox}
\end{center}
\label{tab:disc_rules_examples}
\end{table}

\begin{table}
\caption{Fidelity, number of rules and coverage of characteristic and discriminative rules output by the standard and the modified versions of CEGA using SHAP.}
\begin{center}
\begin{adjustbox}{width=0.95\textwidth}
\small
\begin{tabular}{l c c c c c c@{\hskip 0.125in}c@{\hskip 0.125in}c c c c c c c}
    \toprule
    \multicolumn{1}{c}{\multirow{2}{*}{\bfseries Dataset}} & 
    \multicolumn{6}{c}{\bfseries Characteristic Rules} &
    \multicolumn{6}{c}{\bfseries Discriminative Rules}\\ \cmidrule(lr){2-13}
    \multicolumn{1}{c}{}& Acc.&AUC&F1&\#Rules&Cov.&&&Acc.&AUC&F1&\#Rules&Cov. \\
    \cmidrule(lr){1-13}
    ada & 0.88 & \bf0.92 & 0.83 & 24 & 1.00 &&& 0.85 & 0.90 & 0.79 & 187 & 0.99 \\ \cmidrule(lr){1-13}
    Adult & 0.89 & \bf0.90 & 0.81 & 7 & 0.90 &&& 0.77 & 0.79 & 0.65 & 177 & 1.00\\ \cmidrule(lr){1-13}
    Bank Marketing & 0.90 & \bf0.86 & 0.75 & 2 & 0.89 &&& 0.88 & 0.73 & 0.55 & 54 & 1.00\\ \cmidrule(lr){1-13}
    Blood Transfusion & 0.91 & \bf0.93 & 0.83 & 2 & 0.86 &&& 0.90 & 0.85 & 0.78 & 11 & 1.00\\ \cmidrule(lr){1-13}
    BNG breast-w & 0.96 & 0.99 & 0.95 & 7 & 0.98 &&& 0.95 & 0.99 & 0.95 & 73 & 1.00\\ \cmidrule(lr){1-13}
    BNG tic-tac-toe & 0.75 & \bf0.82 & 0.72 & 10 & 0.99 &&& 0.71 & 0.78 & 0.69 & 13 & 1.00\\ \cmidrule(lr){1-13}
    Compas & 0.88 & 0.75 & 0.67 & 53 & 0.92 &&& 0.75 & \bf0.80 & 0.63 & 511 & 0.76\\ \cmidrule(lr){1-13}
    Churn & 0.89 & \bf0.78 & 0.69 & 4 & 0.91 &&& 0.87 & 0.77 & 0.65 & 224 & 0.99\\ \cmidrule(lr){1-13}
    German Credit & 0.80 & \bf0.82 & 0.69 & 29 & 1.00 &&& 0.78 & 0.81 & 0.62 & 110 & 1.00\\ \cmidrule(lr){1-13}
    Internet Advertisements & 0.96 & 0.92 & 0.91 & 6 & 1.00 &&& 0.95 & \bf0.94 & 0.89 & 24 & 1.00\\ \cmidrule(lr){1-13}
    Jungle Chess 2pcs & 0.70 & \bf0.68 & 0.70 & 3 & 0.89 &&& 0.64 & 0.67 & 0.63 & 73 & 1.00\\ \cmidrule(lr){1-13}
    kc1 & 0.79 & 0.70 & 0.62 & 6 & 0.98 &&& 0.82 & \bf0.72 & 0.61 & 184 & 0.97\\ \cmidrule(lr){1-13}
    mc1 & 0.99 & \bf0.84 & 0.73 & 6 & 1.00 &&& 0.99 & 0.49 & 0.50 & 74 & 1.00\\ \cmidrule(lr){1-13}
    mofn-3-7-10 & 1.0 & 1.0 & 1.0 & 8 & 1.00 &&& 1.0 & 1.0 & 1.0 & 9 & 1.00\\ \cmidrule(lr){1-13}
    Mushroom & 0.97 & 1.0 & 0.97 & 9 & 1.00 &&& 0.95 & 1.0 & 0.95 & 30 & 1.00\\ \cmidrule(lr){1-13}
    Phishing Websites & 0.92 & \bf0.98 & 0.92 & 29 & 1.00 &&& 0.72 & 0.80 & 0.71 & 162 & 1.00\\ \cmidrule(lr){1-13}
    socmob & 0.88 & 0.89 & 0.83 & 5 & 0.76 &&& 0.89 & \bf0.90 & 0.83 & 11 & 0.84\\ \cmidrule(lr){1-13}
    Spambase & 0.91 & \bf0.95 & 0.91 & 15 & 0.81 &&& 0.84 & 0.91 & 0.84 & 88 & 1.00\\ \cmidrule(lr){1-13}
    Steel Plates Fault & 1.0 & \bf1.0 & 1.0 & 9 & 1.00 &&& 0.63 & 0.66 & 0.60 & 136 & 1.00\\ \cmidrule(lr){1-13}
    Telco Customer Churn & 0.85 & \bf0.92 & 0.79 & 4 & 0.74 &&& 0.85 & 0.91 & 0.78 & 51 & 0.99\\ \cmidrule(lr){1-13}
    Average rank & 1.175 & 1.275 & 1.075 & 1 & 1.675 &&& 1.825 & 1.725 & 1.925 & 2 & 1.325\\
    \bottomrule
\end{tabular}
\end{adjustbox}
\end{center}
\label{tab:rulesAcc}
\end{table}

\newpage
\subsection{Comparing Local Explanation Techniques}

Since CEGA is agnostic to the local explanation technique, it is possible to compare the rules obtained using different local explanation techniques, which allows for evaluating the local explainer on a specific task and selecting the one with the highest fidelity. Accordingly, we compare the fidelity, the number of rules and coverage of CEGA when used together with Anchors, SHAP, and LIME, respectively. The result from this comparison is shown in table~\ref{tab:rulesComb}. The results indicate that the characteristic rules produced by CEGA when using Anchors and SHAP tend to provide higher fidelity than when using LIME for most datasets.

\begin{table}
\caption{Fidelity, number of rules and coverage of characteristic rules output by CEGA using Anchors, SHAP, or LIME}
\begin{center}
\begin{adjustbox}{width=1.0\textwidth}
\small
\begin{tabular}{l c c c c c c@{\hskip 0.125in} c@{\hskip 0.125in} c c c c c c@{\hskip 0.125in} c@{\hskip 0.125in} c c c c c}
    \toprule
    \multicolumn{1}{c}{\multirow{2}{*}{\bfseries Dataset}} & 
    \multicolumn{5}{c}{\bfseries Anchors} &
    \multicolumn{9}{c}{\bfseries SHAP} &
    \multicolumn{5}{c}{\bfseries LIME}\\ \cmidrule(lr){2-20}
    \multicolumn{1}{c}{}& Acc.&AUC&F1&\#Rules&Cov.&&&Acc&AUC&F1&\#Rules&Cov.&&&Acc&AUC&F1&\#Rules&Cov. \\
    \cmidrule(lr){1-20}
    ada & 0.87 & 0.91 & 0.81 & 5 & 1.00 &&& 0.88 & \bf0.92 & 0.83 & 24 & 1.00 &&& 0.80 & 0.82 & 0.71 & 27 & 1.00\\ \cmidrule(lr){1-20}
    Adult & 0.88 & \bf0.91 & 0.82 & 7 & 1.00 &&& 0.89 & 0.90 & 0.81 & 7 & 0.90 &&& 0.84 & 0.88 & 0.76 & 47 & 1.00\\ \cmidrule(lr){1-20}
    Bank Marketing & 0.91 & 0.80 & 0.65 & 11 & 1.00 &&& 0.90 & \bf0.86 & 0.75 & 2 & 0.89 &&& 0.87 & 0.68 & 0.55 & 36 & 1.00\\ \cmidrule(lr){1-20}
    Blood Transfusion & 0.93 & 0.92 & 0.86 & 4 & 0.92 &&& 0.91 & \bf0.93 & 0.83 & 2 & 0.86 &&& 0.88 & 0.92 & 0.72 & 5 & 0.8\\ \cmidrule(lr){1-20}
    BNG breast-w & 0.97 & 1.00 & 0.97 & 8 & 0.99 &&& 0.96 & 0.99 & 0.95 & 7 & 0.98 &&& 0.98 & 1.00 & 0.97 & 10 & 0.93\\ \cmidrule(lr){1-20}
    BNG tic-tac-toe & 0.75 & 0.77 & 0.72 & 4 & 1.00 &&& 0.75 & \bf0.82 & 0.72 & 10 & 0.99 &&& 0.73 & 0.76 & 0.69 & 18 & 1.00\\ \cmidrule(lr){1-20}
    Compas & 0.86 & \bf0.82 & 0.67 & 10 & 1.00 &&& 0.88 & 0.75 & 0.67 & 53 & 0.92 &&& 0.87 & 0.74 & 0.64 & 13 & 1.00\\ \cmidrule(lr){1-20}
    Churn & 0.81 & 0.66 & 0.54 & 20 & 1.00 &&& 0.89 & \bf0.78 & 0.69 & 4 & 0.91 &&& 0.87 & 0.66 & 0.58 & 10 & 1.00\\ \cmidrule(lr){1-20}
    German Credit & 0.79 & 0.79 & 0.72 & 6 & 1.00 &&& 0.80 & \bf0.82 & 0.69 & 29 & 1.00 &&& 0.80 & 0.80 & 0.71 & 35 & 1.00\\ \cmidrule(lr){1-20}
    Internet Advertisements & 0.91 & 0.80 & 0.77 & 4 & 1.00 &&& 0.96 & \bf0.92 & 0.91 & 6 & 1.00 &&& 0.81 & 0.73 & 0.67 & 72 & 0.81\\ \cmidrule(lr){1-20}
    Jungle Chess 2pcs & 0.89 & 0.91 & 0.89 & 6 & 1.00 &&& 0.70 & 0.68 & 0.70 & 3 & 0.89 &&& 0.95 & \bf0.96 & 0.95 & 6 & 1.00\\ \cmidrule(lr){1-20}
    kc1 & 0.85 & \bf0.86 & 0.72 & 4 & 0.99 &&& 0.79 & 0.70 & 0.62 & 6 & 0.98 &&& 0.82 & 0.76 & 0.59 & 26 & 1.00\\ \cmidrule(lr){1-20}
    mc1 & 0.97 & \bf0.95 & 0.54 & 19 & 0.99 &&& 0.99 & 0.84 & 0.73 & 6 & 1.00 &&& 0.97 & 0.92 & 0.50 & 13 & 1.00\\ \cmidrule(lr){1-20}
    mofn-3-7-10 & 1.00 & 1.00 & 1.00 & 7 & 0.99 &&& 1.00 & 1.00 & 1.00 & 8 & 1.00 &&& 1.00 & 1.00 & 1.00 & 21 & 1.00\\ \cmidrule(lr){1-20}
    Mushroom & 0.88 & 0.93 & 0.88 & 3 & 1.00 &&& 0.97 & \bf1.00 & 0.97 & 9 & 1.00 &&& 0.93 & 0.97 & 0.93 & 24 & 1.00\\ \cmidrule(lr){1-20}
    Phishing Websites & 0.92 & 0.96 & 0.92 & 5 & 1.00 &&& 0.92 & \bf0.98 & 0.92 & 29 & 1.00 &&& 0.77 & 0.86 & 0.76 & 28 & 1.00\\ \cmidrule(lr){1-20}
    socmob & 0.96 & \bf0.99 & 0.93 & 7 & 1.00 &&& 0.88 & 0.89 & 0.83 & 5 & 0.76 &&& 0.91 & 0.97 & 0.86 & 55 & 1.00\\ \cmidrule(lr){1-20}
    Spambase & 0.90 & \bf0.97 & 0.90 & 8 & 0.93 &&& 0.91 & 0.95 & 0.91 & 15 & 0.81 &&& 0.89 & 0.95 & 0.88 & 34 & 1.00\\ \cmidrule(lr){1-20}
    Steel Plates Fault & 0.77 & 0.82 & 0.77 & 4 & 1.00 &&& 1.00 & \bf1.00 & 1.00 & 9 & 1.00 &&& 0.92 & 0.98 & 0.92 & 12 & 1.00\\ \cmidrule(lr){1-20}
    Telco Customer Churn & 0.89 & \bf0.95 & 0.85 & 8 & 1.00 &&& 0.85 & 0.92 & 0.79 & 4 & 0.74 &&& 0.83 & 0.89 & 0.75 & 27 & 1.00\\ \cmidrule(lr){1-20}
    Average rank & 1.975 & 1.825 & 1.75 & 1.55 & 1.775 &&& 1.675 & 1.775 & 1.725 & 1.675 & 2.425 &&& 2.35 & 2.4 & 2.525 & 2.775 & 1.8\\
    \bottomrule
\end{tabular}
\end{adjustbox}
\end{center}
\label{tab:rulesComb}
\end{table}

\clearpage
\section{Concluding remarks}
\label{CR}

We have proposed CEGA, a method to aggregate local explanations into general characteristic explanatory rules. CEGA is agnostic to the local explanation technique and can work with local explanations in the form of rules or feature scores, given that they are properly converted to itemsets. We have presented results from a large-scale empirical evaluation, comparing CEGA to Anchors and GLocalX, with respect to three fidelity metrics (accuracy, AUC and F1 score), number of rules and coverage. CEGA was observed to significantly outperform GLocalX with respect to fidelity and Anchors with respect to the number of generated rules. We also investigated changing the rule format of CEGA to discriminative rules and using SHAP, LIME, or Anchors as the local explanation technique. The main conclusion of the former investigation is that indeed the rule format has a significant effect; the characteristic rules result in higher fidelity and fewer rules compared to when using discriminative rules. The results from the second follow-up investigation showed that CEGA combined with either SHAP or Anchors generally result in rules with higher fidelity compared to when using LIME as the local explanation technique. 

One direction for future work would be to complement the functionally-grounded (objective) evaluation of the quality of the explanations with user-grounded evaluations, e.g., asking users to indicate whether they actually can follow the logic behind the predictions or solve some tasks using the output rules.

Another direction for future work concerns investigating additional ways of forming itemsets from which general (characteristic or discriminative) rules are formed. This could for example include combining the output of multiple local explanation techniques. Another important direction concerns quantifying the uncertainty of the generated rules, capturing to what extent one can expect a rule to agree with the output of a black-box model. The investigated applications may also include datasets beyond regular tabular data, e.g., text documents and images.

\subsubsection{Acknowledgement}

This work was partially supported by the Wallenberg AI, Autonomous Systems and Software Program (WASP) funded by the Knut and Alice Wallenberg Foundation. HB was partly funded by the Swedish Foundation for Strategic Research (CDA, grant no. BD15-0006).

%
%
%

\begin{thebibliography}{32}
    \bibitem{lime}
    Ribeiro, M., Singh, S., Guestrin, C.: "Why Should I Trust You?": Explaining the Predictions of Any Classifier. { Proceedings Of The 22nd ACM SIGKDD International Conference On Knowledge Discovery And Data Mining, San Francisco, CA, USA, August 13-17, 2016}. pp. 1135-1144 (2016)
    
    \bibitem{shap}
    Lundberg, S., Lee, S.: A Unified Approach to Interpreting Model Predictions. { Advances In Neural Information Processing Systems 30}. pp. 4765-4774 (2017)
    
    \bibitem{anchors}
    Ribeiro, M., Singh, S., Guestrin, C.:  Anchors: High-Precision Model-Agnostic Explanations . { AAAI Conference On Artificial Intelligence (AAAI)}. (2018)
    
    \bibitem{Agrawal:1994}
    Agrawal, R., Srikant, R.: Fast Algorithms for Mining Association Rules in Large Databases. { Proceedings Of The 20th International Conference On Very Large Data Bases}. pp. 487-499 (1994)
    
    \bibitem{Kohavi97}
    Kohavi, R., Becker, B., Sommerfield, D.: Improving Simple Bayes. { European Conference On Machine Learning}. (1997)

    
    \bibitem{Pantelis-AI-review}
    Linardatos, P., Papastefanopoulos, V., Kotsiantis, S.: Explainable AI: A Review of Machine Learning Interpretability Methods. { Entropy}. \textbf{23} (2021)
    
    \bibitem{molnar2019}
    Molnar, C.: Interpretable Machine Learning: A Guide for Making Black Box Models Explainable.  (2019)
    
 
    \bibitem{delaunay:hal-03133223}
    Delaunay, J., Galárraga, L., Largouët, C.: Improving Anchor-based Explanations. { CIKM 2020 - 29th ACM International Conference On Information And Knowledge Management}. pp. 3269-3272 (2020,10)
    
    \bibitem{MAME}
    Natesan Ramamurthy, K., Vinzamuri, B., Zhang, Y., Dhurandhar, A.: Model Agnostic Multilevel Explanations. { Advances In Neural Information Processing Systems}. \textbf{33} pp. 5968-5979 (2020)
    
    \bibitem{GLocalX}
    Setzu, M., Guidotti, R., Monreale, A., Turini, F., Pedreschi, D., Giannotti, F.: GLocalX - From Local to Global Explanations of Black Box AI Models. { Artificial Intelligence}. \textbf{294} pp. 103457 (2021,1)
    
    \bibitem{xgboost}
    Chen, T., Guestrin, C.: XGBoost: A Scalable Tree Boosting System.  (2016,8)
    
    \bibitem{Guidotti_survey}
    Guidotti, R., Monreale, A., Ruggieri, S., Turini, F., Giannotti, F., Pedreschi, D.: A Survey of Methods for Explaining Black Box Models. { ACM Comput. Surv.}. \textbf{51} (2018,8)
    
    
    \bibitem{BostroemGurungLindgren2018_1000117720}
    Boström, H., Gurung, R., Lindgren, T., Johansson, U.: Explaining Random Forest Predictions with Association Rules. { Archives Of Data Science, Series A (Online First)}. \textbf{5}, A05, 20 S. online (2018)
    
    \bibitem{sirus-benard21a}
    Bénard, C., Biau, G., Veiga, S., Scornet, E.:  Interpretable Random Forests via Rule Extraction . { Proceedings Of The 24th International Conference On Artificial Intelligence And Statistics}. \textbf{130} pp. 937-945 (2021,4,13)
    
    \bibitem{FriedmanPopescu}
    Friedman, J., Popescu, B.: Predictive learning via rule ensembles. { The Annals Of Applied Statistics}. \textbf{2} pp. 916-954 (2008)
    
    \bibitem{Model:Agnostic}
    Ribeiro, M., Singh, S., Guestrin, C.:  Model-Agnostic Interpretability of Machine Learning . { ICML Workshop On Human Interpretability In Machine Learning (WHI)}. (2016)
    
    \bibitem{Furnkranz}
    Fürnkranz, J., Kliegr, T., Paulheim, H.: On Cognitive Preferences and the Plausibility of Rule-Based Models. { Mach. Learn.}. \textbf{109}, 853-898 (2020,4)
    
    \bibitem{KLIEGR2021103458}
    Kliegr, T., Bahník, Š., Fürnkranz, J.: A review of possible effects of cognitive biases on interpretation of rule-based machine learning models. { Artificial Intelligence}. \textbf{295} pp. 103458 (2021)
    
    \bibitem{deeplift-shrikumar17a}
    Shrikumar, A., Greenside, P., Kundaje, A.: Learning Important Features Through Propagating Activation Differences. { Proceedings Of The 34th International Conference On Machine Learning}. \textbf{70} pp. 3145-3153 (2017,8,6)
    
    \bibitem{wangCNNExplainerLearning2020}
    Wang, Z., Turko, R., Shaikh, O., Park, H., Das, N., Hohman, F., Kahng, M., Chau, D.: CNN Explainer: Learning Convolutional Neural Networks with Interactive Visualization. { IEEE Transactions On Visualization And Computer Graphics (TVCG)}. (2020)
    
    \bibitem{chr_rule_quan}
    Turmeaux, T., Salleb, A., Vrain, C., Cassard, D.: Learning Characteristic Rules Relying on Quantified Paths. { Knowledge Discovery In Databases: PKDD 2003, 7th European Conference On Principles And Practice Of Knowledge Discovery In Databases, Cavtat-Dubrovnik, Croatia, September 22-26, 2003, Proceedings}. \textbf{2838} pp. 471-482 (2003)
    
    \bibitem{Clark91ruleinduction}Clark, P., Boswell, R.: Rule Induction with CN2: Some Recent Improvements. (Springer-Verlag,1991)

    \bibitem{Cohen95fasteffective}Cohen, W.: Fast Effective Rule Induction. {\em In Proceedings Of The Twelfth International Conference On Machine Learning}. pp. 115-123 (1995)
    
    \bibitem{Friedman1999}Friedman, J., Fisher, N.: Bump hunting in high-dimensional data. {\em Statistics And Computing}. \textbf{9}, 123-143 (1999,4), https://doi.org/10.1023/A:1008894516817
    
    \bibitem{Deng2018InterpretingTE}
    Deng, H.: Interpreting tree ensembles with inTrees. { International Journal Of Data Science And Analytics}. \textbf{7} pp. 277-287 (2018)
    
    \bibitem{Friedman_test}
    Friedman, M.: A Correction: The use of ranks to avoid the assumption of normality implicit in the analysis of variance. { Journal Of The American Statistical Association}. \textbf{34}, 109-109 (1939)
    
    \bibitem{nemenyi1963distribution}
    Nemenyi, P.: Distribution-free multiple comparisons. (Princeton University,1963)

    \bibitem{wilcoxon1945individual}
    Wilcoxon, F.: Individual comparisons by ranking methods. biometrics bulletin 1, 6 (1945), 80–83. {\em URL Http://www. Jstor. Org/stable/3001968}. (1945)
    
    \bibitem{advlime:aies20}
    Slack, D., Hilgard, S., Jia, E., Singh, S., Lakkaraju, H.: Fooling LIME and SHAP: Adversarial Attacks on Post hoc Explanation Methods. { AAAI/ACM Conference On AI, Ethics, And Society (AIES)}. (2020)
    
    \bibitem{Loyola_8882211}
    Loyola-González, O.: Black-Box vs. White-Box: Understanding Their Advantages and Weaknesses From a Practical Point of View. { IEEE Access}. \textbf{7} pp. 154096-154113 (2019)
    
    \bibitem{Furnkranz_rule_learning}Fürnkranz, J., Gamberger, D., Lavrac, N.: Foundations of Rule Learning. (Springer,2012), http://dx.doi.org/10.1007/978-3-540-75197-7
    
    \bibitem{MICHALSKI1983}Michalski, R.: A theory and methodology of inductive learning. {\em Artificial Intelligence}. \textbf{20}, 111-161 (1983), https://www.sciencedirect.com/science/article/pii/0004370283900164
\end{thebibliography}
%

\end{document}